%% file: contextpolicy.tex
\documentclass[letterpaper, 10 pt, conference]{ieeeconf}  

\IEEEoverridecommandlockouts                              
\input{ourcommands}

\title{\LARGE \bf
NVSPolicy: Adaptive Novel-View Synthesis for Generalizable Language-Conditioned Policy Learning
}

\author{Le Shi, Yifei Shi*, Xin Xu*, Tenglong Liu, Junhua Xi, and Chengyuan Chen
\thanks{$\bullet$ The authors are with the College of Intelligence Science and Technology, National University of Defense Technology, China. This work was supported by the National Natural Science Foundation of China (U24A20279), the Natural Science Foundation of Hunan Province (2023JJ20051), the Science and Technology Innovation Program of Hunan Province (2023RC3011), and the Cornerstone Foundation of NUDT (JS24-03).}
\thanks{*Corresponding authors: Yifei Shi (yifei.j.shi@gmail.com) and Xin Xu (xinxu@nudt.edu.cn).}
}

\begin{document}

\maketitle
\thispagestyle{empty}
\pagestyle{empty}

\begin{abstract}
Recent advances in deep generative models demonstrate unprecedented zero-shot generalization capabilities, offering great potential for robot manipulation in unstructured environments.
Given a partial observation of a scene, deep generative models could generate the unseen regions and therefore provide more context, which enhances the capability of robots to generalize across unseen environments.
However, due to the visual artifacts in generated images and inefficient integration of multi-modal features in policy learning, this direction remains an open challenge.
We introduce NVSPolicy, a generalizable language-conditioned policy learning method that couples an adaptive novel-view synthesis module with a hierarchical policy network.
Given an input image, NVSPolicy dynamically selects an informative viewpoint and synthesizes an adaptive novel-view image to enrich the visual context.
To mitigate the impact of the imperfect synthesized images, we adopt a cycle-consistent VAE mechanism that disentangles the visual features into the semantic feature and the remaining feature.
The two features are then fed into the hierarchical policy network respectively: the semantic feature informs the high-level meta-skill selection, and the remaining feature guides low-level action estimation.
Moreover, we propose several practical mechanisms to make the proposed method efficient.
Extensive experiments on CALVIN demonstrate the state-of-the-art performance of our method.
Specifically, it achieves an average success rate of 90.4\% across all tasks, greatly outperforming the recent methods. 
Ablation studies confirm the significance of our adaptive novel-view synthesis paradigm. 
In addition, we evaluate NVSPolicy on a real-world robotic platform to demonstrate its practical applicability.




\end{abstract}

\input{intro}

\input{related}

\input{method}

\input{result}

\input{conclusion}

{\small
\bibliographystyle{ieee_fullname}
\bibliography{egbib}
}

\onecolumn
\clearpage
\input{appendix}

\end{document}

%% file: ourcommands.tex
\usepackage{mathtools}
\usepackage{amsmath}
\usepackage{diagbox}

\usepackage{amssymb}
\usepackage{amsfonts}
\usepackage{amsopn}
\usepackage{graphicx} 
\usepackage{textcomp}
\usepackage{xfrac}
\usepackage{bbm}
\usepackage{overpic}
\usepackage{subfig}
\usepackage{wrapfig}
\usepackage[ruled,vlined,linesnumbered]{algorithm2e}
\usepackage{multirow}
\usepackage{verbatim}
\usepackage{booktabs}
\usepackage{multirow}
\usepackage{array}

\usepackage{color}
\definecolor{green}{rgb}{0, 0.5, 0}
\definecolor{orange}{rgb}{0.8, 0.6, 0.2}
\definecolor{red}{rgb}{1.0, 0.0, 0.0}
\definecolor{teal}{rgb}{0.0, 0.4, 0.4}
\definecolor{purple}{rgb}{0.65,0,0.65}
\definecolor{saffron}{rgb}{0.95,0.75,0.2}
\definecolor{turquoise}{rgb}{0.0,0.5,0.5}
\definecolor{brown}{rgb}{0.5, 0.16, 0.16}

\usepackage{overpic}
\usepackage{currfile} 

\newlength\savedwidth

\definecolor{lightgray}{rgb}{0.6, 0.6, 0.6}

\usepackage[normalem]{ulem}

\usepackage{hyperref}

\newcommand{\hidecomment}[1]{}

\usepackage{xspace}



%% file: intro.tex
\section{Introduction}
\label{sec:intro}

Recent advancements in deep generative models have demonstrated their capability to produce novel content akin to the training data, spanning multiple modalities, including text~\cite{singhal2023large,chowdhery2023palm}, images~\cite{qian20243d,ze20243d}, and videos~\cite{jiang2023vima}.
These models hold significant potential for enhancing robotic manipulation.
Deep generative models could empower robots to generalize across various objects and environments by providing open-world knowledge, thereby circumventing the need for extensive training on large-scale robotic datasets.
This capability is especially advantageous in complex scenarios characterized by high observation incompleteness~\cite{qian20243d} and long-horizon manipulation tasks~\cite{zhou2024language,mees2022matters}.

Given a partial observation of a scene, deep generative models could generate the unseen regions.
This paradigm supplies robots with additional context and, when paired with an adaptive viewpoint-selection strategy, can generation effort on the most informative parts of the workspace.
This paradigm provides robots with more context, bringing generalizable open-world knowledge.
Despite the advantages, incorporating deep generative models with policy learning remains challenging in the field.
One main reason is that the generated content is not always faithful and might contain substantial visual artifacts. As such, simply relying on the generated content might degrade the robot's performance.
On the other hand, the aggregation of multi-modal features in policy learning is a long-standing problem.
How to generate informative and reliable multi-modal features from less faithful observations requires careful network designs.

In this paper, we leverage adaptive novel-view synthesis to enhance generalizable language-conditioned policy learning.
Unlike conventional generative models that produce panoramas or full 3D scene reconstructions, our approach synthesizes novel images from previously unseen viewpoints.
These viewpoints are adaptively selected based on the robot–object spatial relationship, providing more informative observations for downstream manipulation.
Novel-view  synthesis is a relatively well-studied task, enabling us to leverage a substantial body of existing research~\cite{seo2024genwarp1,kong2024eschernet}.
By incorporating appropriate viewpoint selection strategies, novel-view synthesis could generate content with high certainty, thereby enhancing the robustness of downstream robotic manipulation tasks.

\begin{figure}[t]
\centering
    \centering
    \includegraphics[width=1.0\linewidth]{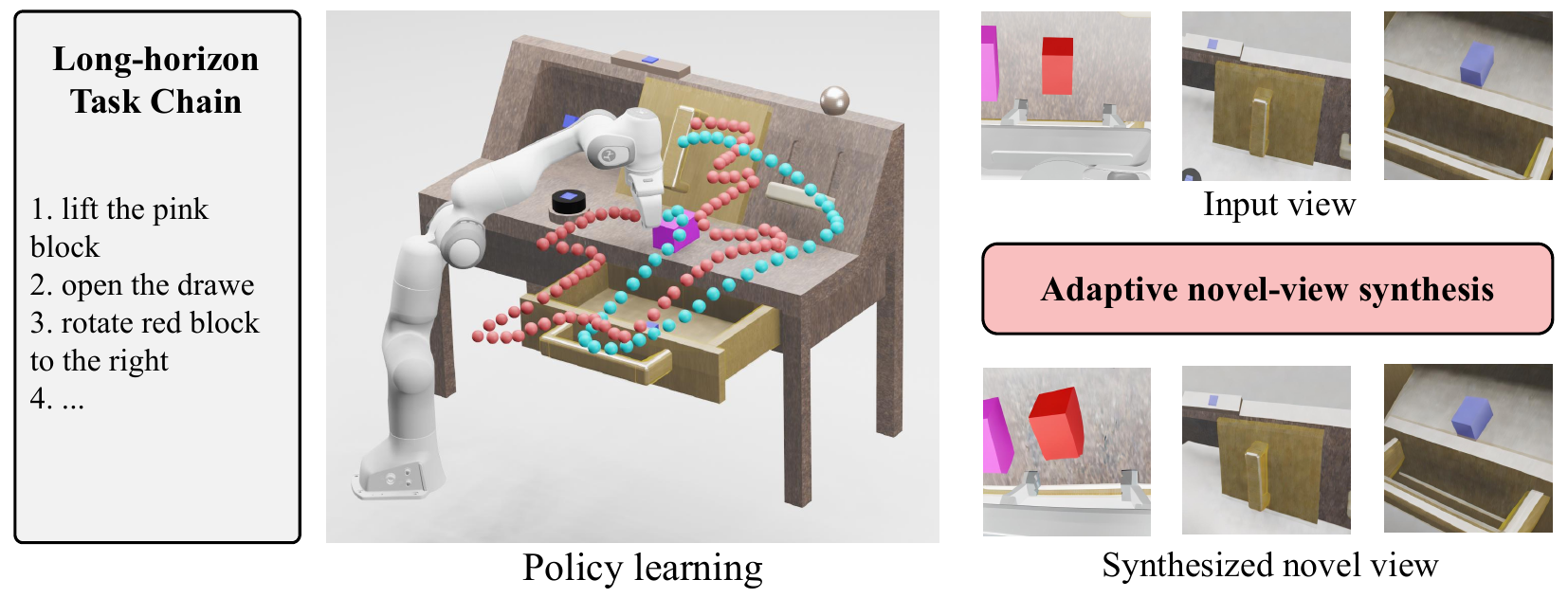}
    \caption{The proposed NVSPolicy leverages adaptive novel-view synthesis to provide more context, boosting the performance of language-conditioned policy learning.}
    \label{fig:teaser}
\end{figure}

To this end, we propose NVSPolicy, a language-conditioned policy learning method that integrates an adaptive novel-view synthesis model with a hierarchical policy network (Fig.~\ref{fig:teaser}).
The method consists of several key components.
\emph{First}, given an input image, it generates context-enhanced observations by synthesizing novel-view images at a adaptive novel viewpoint near the robot's current perspective.
\emph{Second}, to mitigate the impact of the imperfect synthesized images, we adopt a cycle-consistent VAE mechanism that disentangles the visual features into the semantic feature and the remaining feature.
\emph{Third}, a hierarchical policy network is utilized to estimate the robot's action.
Specifically, it first selects the high-level meta-skill with the semantic feature. It then generates the low-level action with the remaining feature, given the selected meta-skills.
\emph{Last}, we propose several practical mechanisms to make the method efficient.

On the challenging CALVIN language-conditioned robotics benchmark, our approach achieves an average success rate of 90.4\% across all tasks and demonstrates a mean completion capacity of 2.93 in five consecutive long-horizon tasks. Moreover, ablation studies underscore the importance of our proposed practical considerations. 
We further evaluate NVSPolicy on a real-world robot to demonstrate its practicality.
We show that NVSPolicy greatly benefits from the adaptive novel-view synthesis to improve adaptability while maintaining good efficiency.

%% file: related.tex
\section{Related Work}
\label{sec:related}

\subsection{Language-conditioned Robotic Manipulation}
Language-conditioned manipulation has advanced from cross-modal attention that tightly aligns vision and language signals for fine-grained object grounding~\cite{ito2022integrated,Xu_Zhao_Zhou_Li_Pi_Zhu_Wang_Xiong_2023} to large vision–language–action models that transfer web-scale knowledge into robotic policies with impressive zero-shot generalization~\cite{jin2023alphablock,xu2024rt}. Recent work further improves instruction comprehension through chain-of-thought reasoning and affordance grounding, enabling robots to follow multi-step, constraint-rich commands~\cite{wei2022chain,vemprala2024chatgpt}. Despite these gains, most methods depend on task-specific demonstrations, limiting scalability. Hierarchical skill learning mitigates this by reusing low-level primitives: GCBC mines unlabeled play to compose goal-conditioned behaviors~\cite{lynch2020language}, and HULC maps language to latent motor skills~\cite{mees2022matters}. SPIL augments imitation with base-skill priors for unstructured scenes but processes raw observations holistically, risking feature interference~\cite{zhou2024language}. 

\subsection{Generative Models for Robot Manipulation}
The development of generative models for robotic manipulation has progressed along two complementary but distinct directions: visual-motor learning and vision-language-action integration. Early breakthroughs in visuomotor control~\cite{levine2018learning, kalashnikov2018scalable} established data-driven approaches for manipulation tasks, while subsequent work on diffusion models~\cite{janner2022planning, zhanglanguage} demonstrated their effectiveness in generating diverse action distributions. However, these methods typically require either extensive demonstration data or suffer from computationally expensive iterative inference processes.
Parallel advances in vision-language foundation models introduced new capabilities for instruction-following manipulation. RT-2~\cite{zitkovich2023rt} pioneered the integration of web-scale vision-language pretraining with robotic control. Subsequent work like RoboFlamingo~\cite{livision} improved generalization through autoregressive architectures, but inherited similar efficiency challenges. Meanwhile, analyses of general vision-language models like VILA~\cite{hu2024look} revealed fundamental limitations in motion-aware grounding that persist across these approaches.

\subsection{Novel-View Synthesis}
Novel-view synthesis (NVS) now follows several main directions. Implicit radiance-field models such as NeRF~\cite{mildenhall2020nerf} and its fast variants~\cite{muller2022instant, chen2023uv} yield photorealistic views but remain too slow for on-board robotics. Explicit 3D Gaussian Splatting (3DGS) replaces ray-marching with rasterisation, reaching real-time rates; recent work compresses the representation and even extends it to dynamic scenes without sacrificing speed~\cite{kerbl20233d, niedermayr2024compressed}. Parallel to these geometry-aware approaches, generative image models such as PI-GAN~\cite{chan2021pi} and NVS-GAN~\cite{shrisha2024nvs}, treat NVS as unconditional synthesis but can suffer mode collapse, while the diffusion-based GenWarp~\cite{seo2024genwarp1} alleviates these issues by introducing semantic-preserving diffusion-based warping, but its fidelity still degrades under large viewpoint shifts. Affine-correspondence–based pose solvers~\cite{Guan_TCYB, Guan_IJCV} derive lightweight geometric priors for 3-D perception from only a few feature matches, thereby offering an efficient foundation for novel-view synthesis. We therefore couple GenWarp with an adaptive viewpoint selector that maximises context gain under robot–object spatial constraints, significantly boosting the success of our language-conditioned policy.

%% file: method.tex
\section{Method}
\label{sec:method}
\maketitle

\begin{figure*}[htp]
\centering
    \centering
    \includegraphics[width=1\linewidth]{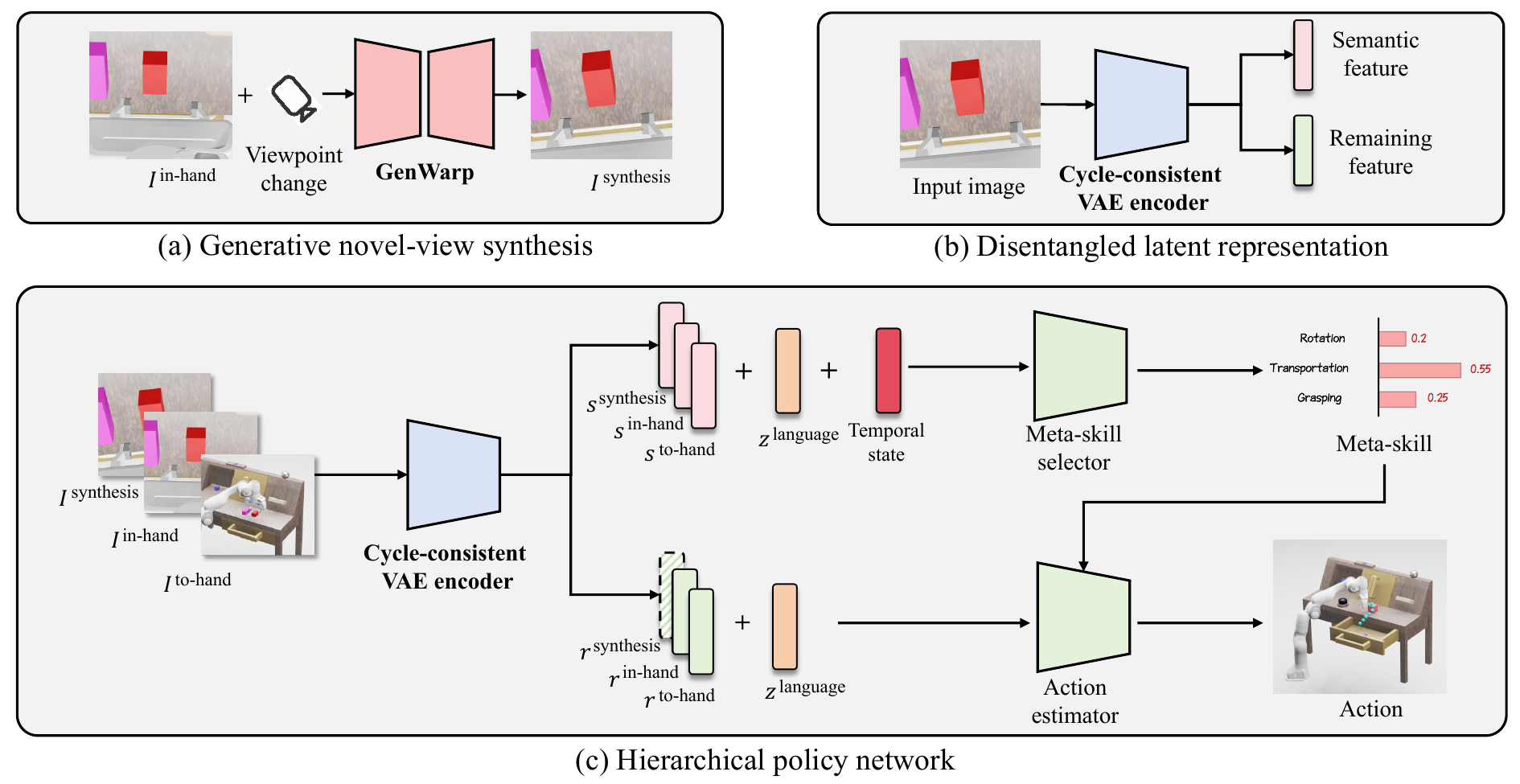}
    \caption{The overview of the proposed NVSPolicy. (a) It synthesizes context-enhanced novel-view images given the input images. (b) It disentangles the image into the semantic feature and the remaining feature with a cycle-consistent VAE mechanism, mitigating the impact of the imperfect synthesized images. (c) Based on (a) and (b), a hierarchical policy network is developed to estimate the optimal meta-skill and the low-level action.}
    \label{fig:overview}
\end{figure*}

\subsection{Overview}\label{overview}
In this work, we focus on long-horizon, language-conditioned manipulation tasks using a robotic arm equipped with a parallel gripper.
An eye-in-hand camera and a fixed eye-to-hand camera are equipped to scan RGB images. At each time step, the method takes the language instruction $l$, the RGB images captured by the eye-in-hand camera $I^{\text{in-hand}}$ and the fixed eye-to-hand camera $I^{\text{to-hand}}$ as inputs. It outputs an action $x = \pi(l, I^{\text{in-hand}}, I^{\text{to-hand}})$, where $\pi(\cdot)$ is the policy.

The proposed NVSPolicy consists of several components.
\emph{First}, it synthesizes context-enhanced novel-view images from adaptively selected viewpoints based on the robot–scene configuration (Sec.~\ref{sec:nvs}).
\emph{Second}, it disentangles the images into the semantic feature and the remaining feature, mitigating the impact of the imperfect synthesized images (Sec.~\ref{sec:latent}).
\emph{Third}, a hierarchical policy network is developed to estimate the robot's action (Sec.~\ref{sec:policy}).
\emph{Last}, several practical mechanisms are introduced to make the policy network efficient during the inference (Sec.~\ref{sec:consider}). Fig.~\ref{fig:overview} illustrates the overview of NVSPolicy.

\subsection{Adaptive Novel-view Synthesis}\label{sec:nvs}

Understanding the unseen region plays a pivotal role in enabling robots to interact with the environment.
To enhance the capabilities of context understanding, we leverage a pre-trained generative model, GenWarp~\cite{seo2024genwarp1}, to synthesize new images from adaptively chosen novel viewpoints.
Specifically, GenWarp generates novel-view images from a single input image by jointly learning to warp the visible content and hallucinate the unseen regions. The model could effectively preserve the semantic structure of the input image while adapting to viewpoint shifts. This leads to high-fidelity novel-view images with strong generalization to out-of-distribution scenes.

When applying GenWarp to our framework, we found that the quality of the generated novel-view images is highly sensitive to the magnitude of camera viewpoint changes.
The generated novel-view images are prone to be less faithful when it comes to large camera viewpoint changes.
To maximize the information gain from the generated images while keeping them faithful, the following viewpoint selection strategy is developed.

\begin{figure}[t!]
\centering
    \centering
    \includegraphics[width=1.0\linewidth]{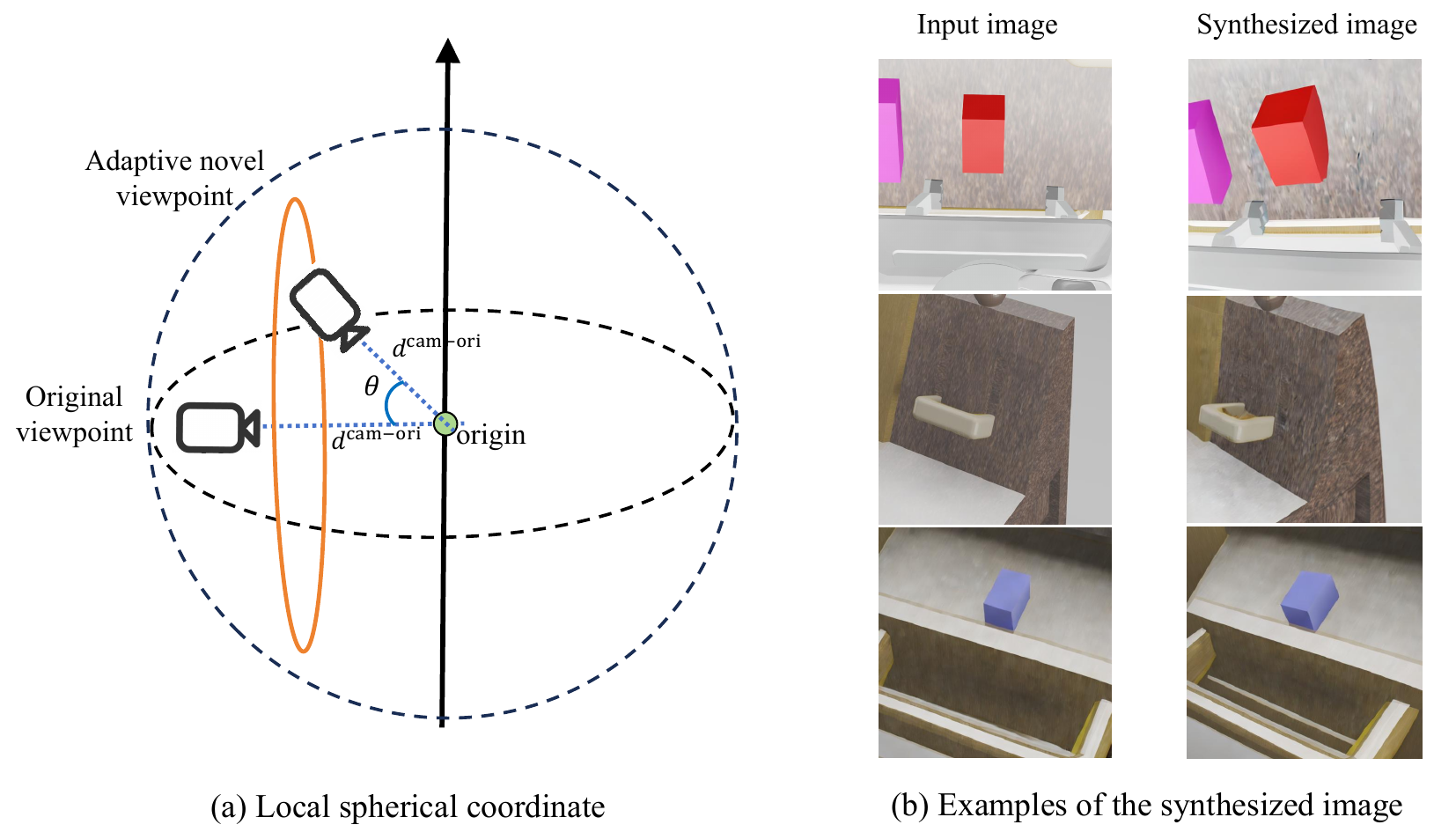}
    \caption{(a) The local spherical coordinate in the adaptive novel viewpoint selection. (b) Examples of the synthesized image.}
    \label{fig:genwarp}
\end{figure}

We first build a local spherical coordinate, as shown in Fig.~\ref{fig:genwarp} (a). The origin of the spherical coordinate system lies in the primary ray of the camera. The distance between the camera and the origin $d^{\text{cam-ori}}$ is equal to the average depth of the input image. The average depth is estimated by the method in~\cite{seo2024genwarp1}.
In our adaptive viewpoint selection scheme, we sample a novel viewpoint whose angular offset $\theta$ is determined based on the average scene depth, ensuring both informative content and stable generation.
$\theta$ is determined by considering $d^{\text{cam-ori}}$:
\begin{equation}
\theta = -w_1 \cdot d^{\text{cam-ori}} + w_2,
\label{eq:viewpoint}
\end{equation}
where $w_1$ and $w_2$ are the hyper-parameters. We use $w_1=14$ and $w_2=39$ in our experiments.
To further increase the randomness, a small location perturbation is added to the sampled viewpoint.
Empirical results show that this adaptive selection strategy consistently yields high-quality synthesized views that preserve semantic coherence while capturing novel, task-relevant content. Examples of the synthesized images are shown in Fig.~\ref{fig:genwarp} (b).

\begin{figure*}[t!]
\centering
    \centering
    \includegraphics[width=1.0\linewidth]{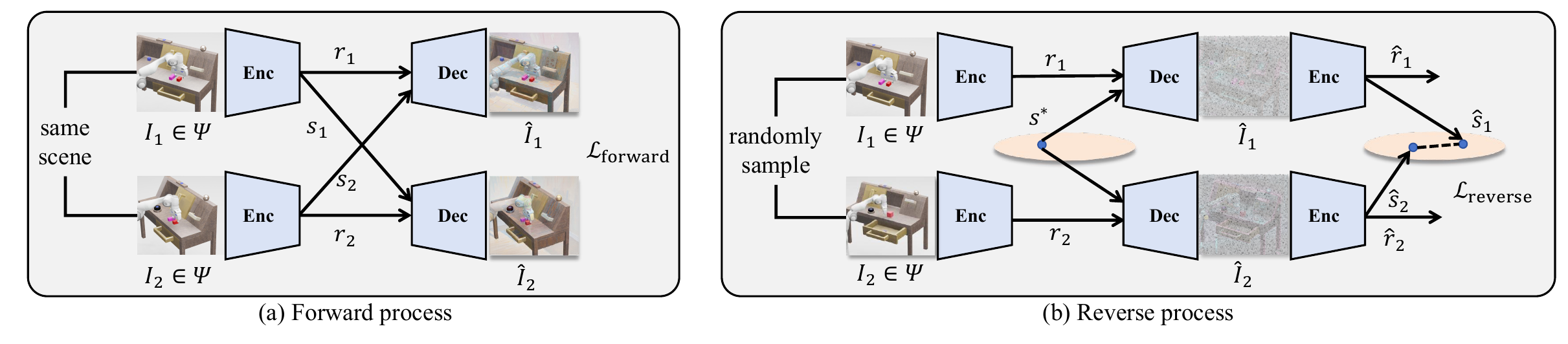}
    \caption{The training protocol of the cycle-consistent VAE: a forward process and a reverse process are developed to jointly optimize the encoder and decoder. Once trained, the encoder can be applied to disentangle the visual features into the semantic feature and the remaining feature during inference.}
    \label{fig:cycle}
\end{figure*}

\subsection{Disentangled Latent Representation}\label{sec:latent}

The synthesized images described above are mostly faithful in terms of the overall structure and semantics.
However, they are still not perfect as local geometric distortions and unrealistic appearances might exist, which would hinder the performance of policy learning.
To mitigate these, we opt to disentangle the visual feature of the synthesized images into the semantic feature and the remaining feature using a cycle-consistent VAE mechansim~\cite{jha2018disentangling}.
The disentangled semantic feature is then fed into a hierarchical policy network to estimate the high-level meta-skill of the robot's action.

To achieve this, we first collect a set of image pairs $\Psi$, where each pair contains two images captured from the same scene with different viewpoints. We also add random local geometric distortions and appearance changes to both images to mimic the imperfection of the synthesized novel-view images.
The training of the cycle-consistent VAE includes a forward process and a reverse process.

In the forward process, as illustrated in Fig.~\ref{fig:cycle} (a), we take each image pair $\psi=\{I_1,I_2\}$ in $\Psi$ as input.
A cycle-consistent VAE encoder $f_{\phi}(\cdot)$, consisting of three convolutional layers and two fully connected layers, is applied to estimate two latent variables for each image:
$(s_1,r_1) = f_\phi (I_1)$, $(s_2,r_2) = f_{\phi}(I_2)$,
where $s_1$ and $s_2$ are the semantic latent variables and $r_1$ and $r_2$ are the remaining latent variables.
We subsequently reconstruct the two images using a cycle-consistent VAE decoder $f_{\omega}(\cdot)$, which mirrors the encoder architecture and comprises two fully connected layers followed by three convolutional layers. The decoder takes as input the exchanged semantic latent variables and the original remaining latent variables to generate reconstructed images: $\hat{I}_1 = f_{\omega}(s_2, r_1)$ and $\hat{I}_2 = f_{\omega}(s_1, r_2)$.
The loss function of the forward process is:
\begin{equation}
\begin{aligned}
\mathcal{L}_{\mathrm{forward}} =&
-\,\mathbb{E}_{q_\phi(s_2 \mid I_2),\, q_\phi(r_1 \mid I_1)}
\bigl[\log\,p_\omega \bigl(I_1 \mid s_2,r_1\bigr)\bigr] \\
-&\mathbb{E}_{q_\phi(s_1 \mid I_1),\, q_\phi(r_2 \mid I_2)}
\bigl[\log\,p_\omega \bigl(I_2 \mid s_1,r_2\bigr)\bigr] \\
+\mathrm{KL}&\bigl(q_\phi\bigl(r_1 \mid I_1\bigr)\|p(r)\bigr)
+\mathrm{KL}\bigl(q_\phi\bigl(r_2 \mid I_2\bigr)\|p(r)\bigr)\\
+\mathrm{KL}&\bigl(q_\phi\bigl(s_1 \mid I_1\bigr)\|p(s)\bigr)
+\mathrm{KL}\bigl(q_\phi\bigl(s_2 \mid I_2\bigr)\|p(s)\bigr),
\end{aligned}
\label{eq:forward_loss}
\end{equation}
where $p_{\omega}(\cdot)$ and $q_{\phi}(\cdot)$ denote the parameterized cycle-consistent VAE decoder and encoder, respectively. The prior distributions $ p(s)$ and $ p(r) $ for the semantic latent variable $ s $ and the remaining latent variable $r$, respectively, are assumed to follow the standard Gaussian distribution.
 $\mathrm{KL}(\cdot)$ represents the Kullback--Leibler divergence.
The expectation operator $\mathbb{E}$ is computed as the expected reconstruction error under the approximate posterior distribution of the latent variables. It is introduced to account for the stochastic nature of the latent representations during training.

In the reverse process, as shown in Fig.~\ref{fig:cycle} (b), we randomly sample two images from $\Psi$ without requiring the scene consistency and feed them through the cycle-consistent VAE encoder to generate the remaining latent variables $r_1$ and $r_2$.
A semantic latent variable $s^{*}$ is sampled from the standard Gaussian distribution.
Then, the cycle-consistent VAE  decoder reconstructs the two images $\hat{I_1}$ and $\hat{I_2}$ with $s^{*}$:
$\hat{I}_1 = f_\omega( s^*, r_1), \, \hat{I}_2 = f_\omega(s^*, r_2)$.
Since $\hat{I_1}$ and $\hat{I_2}$ are reconstructed with the same semantic latent variable, the estimated semantic latent variable from $\hat{I_1}$ and $\hat{I_2}$ should be identical.
To enforce this constraint, we re-encode the generated images and compute their semantic embedding:
$\hat{s}_1 = f_\phi(\hat{I}_1), \,
\hat{s}_2 =  f_\phi(\hat{I}_2)$.
The loss function of the reverse process is:
\begin{equation}
\mathcal{L}_{\text{reverse}} =
\mathbb{E}_{s^* \sim p(s)}
(
\left\|\, \hat{s}_1 - \hat{s}_2 \,\right\|_1
),
\label{eq:reverse_loss}
\end{equation}
where the expectation operator $\mathbb{E}_{s^* \sim p(s)}$ denotes the expected reconstruction consistency error computed over latent semantic variables $ s^* $ sampled from the standard Gaussian prior $p(s)$.

The cycle-consistent VAE is trained using the overall loss function: $\mathcal{L}=\mathcal{L}_\text{forward}+ \lambda \cdot \mathcal{L}_\text{reverse}$, where $\lambda=0.5$ is the hyper-parameter.
Once trained, we can adopt the cycle-consistent VAE encoder to disentangle the visual features into the semantic feature and the remaining feature during inference.

\begin{table*}[!ht]
\caption{The performance of NVSPolicy and competing baselines on the CALVIN benchmark.
}
\label{method-baselin-result}
\vskip 0.05in
\centering
\small
\setlength{\tabcolsep}{9pt}
\begin{tabular}{cccccc||c}
\hline
\toprule
\diagbox{\textbf{Horizon}}{\textbf{Method}}
& \multicolumn{1}{c}{\bf  GCBC~\cite{lynch2020language}} & \multicolumn{1}{c}{\bf MCIL~\cite{mees2022calvin}} & \multicolumn{1}{c}{\bf HULC~\cite{mees2022matters}} & \multicolumn{1}{c}{\bf SPIL~\cite{zhou2024language}} & \multicolumn{1}{c}{\bf LCD~\cite{zhanglanguage}} &
\multicolumn{1}{c}{\bf Ours} \\
\midrule
    1 & $64.7 \pm 4.0 $ & $76.4 \pm 1.5$ & $82.6 \pm 2.6$ & $84.6 \pm 0.6$ & $88.7 \pm 1.5$ & $\mathbf{90.4 \pm 0.3}$\\
2 & $ 28.4 \pm 6.2 $ & $48.8 \pm 4.1$ & $64.6 \pm 2.7$ & $65.1 \pm 1.3$ & $69.9 \pm 2.8$ & $\mathbf{70.4 \pm 1.2}$\\
3 & $12.2 \pm 4.1$ & $30.1 \pm 4.5$ & $47.6 \pm 3.2$ & $50.8 \pm 0.4$ & $51.5 \pm 5.0$& $\mathbf{56.3 \pm 1.5}$\\
4 & $4.9 \pm 2.0$ & $18.1 \pm 3.0$ & $36.4 \pm 2.4$ & $38.0 \pm 0.6$ & $42.7 \pm 5.2$& $\mathbf{44.1 \pm 0.6}$\\
5 & $1.3 \pm 0.9 $ & $9.3\pm3.5$ & $26.5 \pm 1.9$ &$28.6 \pm 0.3$  &$30.2 \pm 5.2$ & $\mathbf{32.8 \pm 0.2}$\\
\midrule
$\text{Avg  horizon  len}$ & $1.11 \pm 0.3$ & $1.82 \pm 0.2$ & $2.57 \pm 0.12$ & $2.67 \pm 0.01$ & $2.88 \pm 0.19$ &$\mathbf{2.93 \pm 0.04}$\\
\bottomrule
\hline
\end{tabular}
\end{table*}

\subsection{Hierarchical Policy Network}\label{sec:policy}
The disentangled latent representation is inherently suitable for hierarchical skill learning: the semantic feature informs
the high-level meta-skill selection, and the remaining feature guides low-level action estimation.
For example, determining the meta-skill (e.g., move or grasp) requires semantic features rather than appearance features. To this end, we propose a hierarchical policy network that first selects the optimal meta-skill and then estimates the action.


The hierarchical policy network takes the images from three viewpoints, along with a language instruction as input. The three images are those captured from a fixed external camera $I^{\text{to-hand}}$, captured from a wrist-mounted gripper camera $I^{\text{in-hand}}$, and a synthesized image from novel viewpoint $I^{\text{synthesis}}$ derived from $I^{\text{in-hand}}$. 
The language instruction is fed into a pre-trained MiniLM~\cite{wang2020minilm}, generating a feature embedding $z^{\text{language}}$. 
At each timestep, the policy network estimates action $a$, where $a \in \mathbb{R}^7$ represents a 7-DoF relative control vector comprising 3D position displacement, 3D orientation change, and a 1D gripper command.

The network starts from extracting features of $I^{\text{to-hand}}$, $I^{\text{in-hand}}$, and $I^{\text{synthesis}}$.
Specifically, we feed them into the cycle-consistent VAE encoder respectively, resulting in $(s^{\text{to-hand}},r^{\text{to-hand}})$, $(s^{\text{in-hand}},r^{\text{in-hand}})$, and $(s^{\text{synthesis}},r^{\text{synthesis}})$.
Moreover, to take advantage of temporal consistency in estimating meta-skills, we employ a rolling buffer of length $T=15$ to maintain the semantic feature of the previous timesteps. This temporal state provides more context, improving the robustness.
Then, a meta-skill selector $q_\xi(o|\cdot)$, similar to~\cite{zhou2024language}, is adopted to aggregate feature from the extracted semantics features $s^{\text{to-hand}}$, $s^{\text{in-hand}}$, $s^{\text{synthesis}}$, the language feature $z^{\text{language}}$, and the temporal state.
The output of the meta-skill selector is a categorical distribution $o \in \mathbb{R}^3$ over the predefined meta-skills.
We adopt the rule-based labeling procedure $p(o|a)$ from~\cite{zhou2024language} to acquire the meta-skill ground truth to train the meta-skill selector.

Then, the selected optimal meta-skill $o^{\text{optimal}}$ is fed into an action estimator to generate the robot's action.
Specifically, an action estimator takes the meta-skill label $o^{\text{optimal}}$, the extracted remaining feature $r^{\text{to-hand}}$, $r^{\text{in-hand}}$, and the language feature $z^{\text{language}}$ as input.
Note that we does not consider $r^{\text{synthesis}}$ as it contains the remaining feature from the synthesized image, which is less reliable.
The action estimator $f_\alpha$ consists of a two-layer Transformer with eight attention heads, similar to~\cite{mees2022matters}.
During training, imitation learning is performed using demonstration actions as supervision. To improve stability, we apply a regularization term as in~\cite{zhou2024language}.
\begin{equation}
\begin{aligned}
    \mathcal{L}_{\text{hierarchical}} &=  {\mathrm{KL}}(a^* \|f_\alpha(o^{\text{optimal}},r^{\text{to-hand}}, r^{\text{in-hand}}, z^{\text{language}}))\\
    + &\gamma_1\mathbb{E}\left[-\sum_{o}p(o|a)\log q_{\xi}(o|s^{\text{all}}, z^{\text{language}})\right]\\
    +&\gamma_2\text{KL}(q_{\xi}(o|s^{\text{all}}, z^{\text{language}})||p(o)),
\end{aligned}
\end{equation}
where $s^{\text{all}}$ denotes semantic features from all three views, $\gamma_1=0.2,\gamma_2=0.01$ are the hyper-parameters. The first term minimizes the KL divergence between the demonstrated action $a^*$ and the predicted action. The second term is a cross-entropy loss that supervises meta-skill prediction using rule-based labels $p(o|a)$. The third term is a regularization that encourages $q_{\xi}(o|\cdot)$ to stay close to the uniform prior distribution $p(o)$, ensuring balanced skill selection.

\subsection{Practical Considerations}\label{sec:consider}
\subsubsection{Keyframe Selection}
Since the generative novel-view synthesis model entails substantial computational costs, performing novel-view synthesis at every timestep is neither efficient nor necessary. To mitigate this, during network training, we select keyframes from consecutive frames and perform novel-view synthesis only on these keyframes.
Specifically, we first designate the initial frame captured by the wrist-mounted gripper camera as a keyframe. Then, we traverse the subsequent frames, computing the viewpoint change relative to the previous keyframe for each frame. A new keyframe is assigned when the viewpoint change exceeds a predefined threshold.
We substitute the synthesized image with a blank image for non-keyframes in the hierarchical policy network.

\subsubsection{Policy Distillation}
To further enable efficient inference of the hierarchy policy network, we introduce a policy distillation module.
The student model does not generate the novel-view images explicitly. Instead, it distills the knowledge from the teacher model by adopting an extra network to estimate the semantic feature of the generated novel-view images by the teacher model.
The policy distillation is optimized by the Kullback–Leibler divergence loss function between the two features: 
$
\mathcal{L}_{\text{PD}} = {\mathrm{KL}}(s^\text{student} \,\|\, s^\text{teacher}),
$
where $s^\text{student}$ denotes the predicted semantic feature by the student model, and $s^\text{teacher}$ is the semantic feature of the synthesized image generated by the cycle-consistent VAE encoder. Due to its low computational cost, the student model enables semantic estimation even on non-key frames, leading to smoother and more consistent policy behavior throughout the task sequence.

%% file: result.tex
\section{Results and Evaluation}
\label{sec:result}

\subsection{Experimental Benchmark}
We conduct evaluations on the CALVIN benchmark~\cite{mees2022calvin}, a large-scale, multi-task dataset requiring agents to complete five language-conditioned subtasks from raw RGB observations and robot actions sequentially.
CALVIN provides $1,000$ task sequences derived from $34$ distinct subtasks, with each episode initialized to a standardized robot configuration to eliminate spatial bias.

\begin{figure*}[thp!]
\centering
    \centering
    \includegraphics[width=1\linewidth]{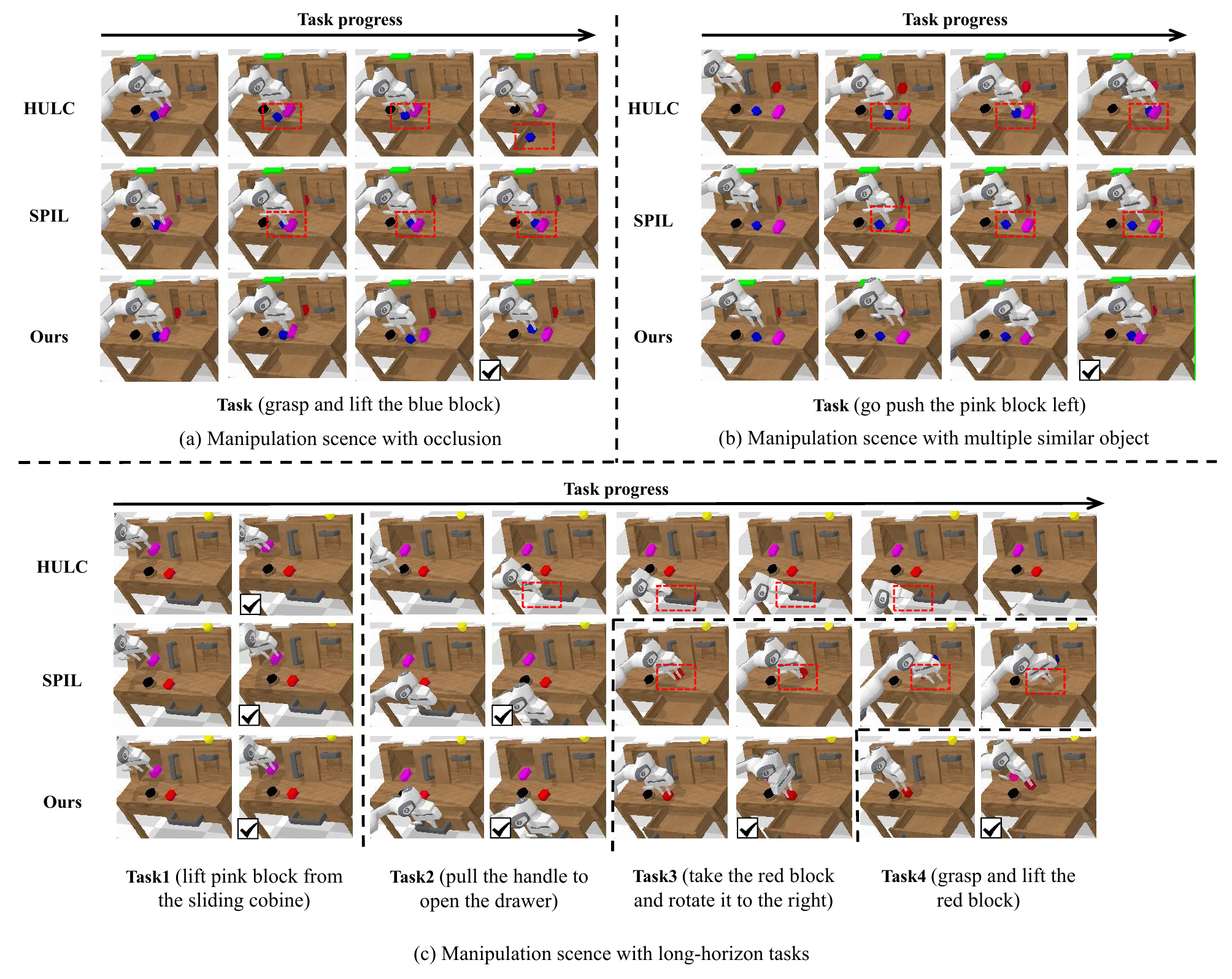}
    \caption{Qualitative comparisons of NVSPolicy with several recent methods. Frames marked with a check mark indicate a successful task completion, while the representative failure cases are highlighted with red dashed boxes.}
    \label{fig:result}
\end{figure*}

\subsection{Evaluation Metrics}
We evaluate NVSPolicy using two evaluation metrics: 1) The task success rates are computed over sequences containing up to five consecutive subtasks. In the Horizon axis, the numbers 1, 2, 3, 4, and 5 represent the success rates for completing 1, 2, 3, 4, and 5 consecutive tasks, respectively. 2) The average successful sequence length is computed as the mean number of consecutive successful tasks. The success rate captures policy effectiveness at varying sequence lengths, while the horizon length reflects robustness in executing long-horizon, language-conditioned task chains. 

\subsection{Compare to Recent Methods}
We compare NVSPolicy to the state-of-the-art methods from the CALVIN leaderboard\footnote{Leaderboard: \url{https://calvin.cs.uni-freiburg.de}}. These include: 1) GCBC~\cite{lynch2020language}, a simple behavior cloning approach without latent structure; 2) MCIL~\cite{mees2022calvin}, which employs a sequential CVAE to model reusable state trajectories for action prediction; 3) HULC~\cite{mees2022matters} leverages hierarchical framework for language-conditioned control; 4) SPIL~\cite{zhou2024language} incorporates skill priors to enhance generalization; 5) LCD~\cite{zhanglanguage} leverages diffusion models across temporal, state, and task spaces for long-horizon, language-conditioned control.
Note that, we only compare to methods without involving the pretrained large-scale language or vision-language foundation models.

\begin{table*}[!t]
\caption{The performance of different novel viewpoint generated images on the CALVIN benchmark.
}
\label{method-angel-result}
\vskip 0.05in
\centering
\small
\setlength{\tabcolsep}{10.5pt}
\begin{tabular}{cccccc||c}
\hline
\toprule
\diagbox{\textbf{Horizon}}{$\mathbf{\theta}$}& \multicolumn{1}{c}{\textbf {$0^\circ$}} & \multicolumn{1}{c}{\textbf {$10^\circ$}}&
\multicolumn{1}{c}{\textbf {$20^\circ$}} & \multicolumn{1}{c}{\textbf {$30^\circ$}} & \multicolumn{1}{c}{\textbf {$40^\circ$}}  &
\multicolumn{1}{c}{\bf Ours} \\
\midrule
1 & $89.1 \pm 0.7 $ & $ 89.5 \pm 1.2$ & $90.1 \pm 0.3$ & $87.4 \pm 0.5$ & $87.1 \pm 2.1 $&$\mathbf{90.4 \pm 0.3}$\\
2 & $ 70.0 \pm 0.3 $ &  $70.8\pm 2.7$& $\mathbf{71.2 \pm 1.8}$ &$ 68.0 \pm 3.8$ & $66.9 \pm 1.3$ & $70.4 \pm 1.2$ \\
3 & $ 51.5 \pm 0.7$ & $ 52.4\pm 2.2$& $54.5 \pm 2.3$ &$50.4 \pm 4.1$  & $49.0 \pm 1.1$ & $\mathbf{56.3 \pm 1.5}$\\
4 & $ 37.7 \pm 1.2 $ & $37.4 \pm 1.8$ & $42.1 \pm 1.2$ &$37.5 \pm 2.6$  & $34.7 \pm 2.1$& $\mathbf{44.1 \pm 0.6}$\\
5 & $ 26.5 \pm 0.6 $ & $26.2 \pm 1.9$& $29.2 \pm 0.6$ & $26.0\pm1.2$ & $23.7 \pm 0.7$  & $\mathbf{32.8 \pm 0.2}$\\
\midrule
$\text{Avg  horizon  len}$ & $2.75 \pm 0.03$ & $2.81\pm 0.05$&$2.89 \pm 0.06$& $2.72 \pm 0.06$  & $2.61 \pm 0.11$ &$\mathbf{2.93 \pm 0.04}$\\
\bottomrule
\hline
\end{tabular}
\end{table*}

\begin{table}[!t]
\caption{Ablation studies. NS represents novel-view synthesis. FD represents feature disentanglement. PD represents policy distillation. 
}
\label{method-result}
\vskip 0.05in
\centering
\small
\setlength{\tabcolsep}{3pt}
\begin{tabular}{cccc||c}
\hline
\toprule
\diagbox{\tiny \textbf{Horizon}}{\tiny\textbf{Method}}& \multicolumn{1}{c}{\bf  w/o NS}& \multicolumn{1}{c}{\bf w/o FD} & \multicolumn{1}{c}{\bf w/o PD} &
\multicolumn{1}{c}{\bf Ours} \\
\midrule
1 &$88.3 \pm 1.1$ & $88.3 \pm 1.1$&  $89.2 \pm 1.6$  & $\mathbf{90.4 \pm 0.3}$\\
2 &$68.7 \pm 1.6$ & $68.7 \pm 1.6$  &  $\mathbf{71.9 \pm 2.2}$ & $70.4 \pm 1.2$\\
3 &$51.3 \pm 2.7$ & $51.3 \pm 2.7$  & $53.9 \pm 3.8$ & $\mathbf{56.3 \pm 1.5}$\\
4 &$36.7 \pm 2.4$ & $36.4 \pm 1.4$& $41.1 \pm 2.3$ & $\mathbf{44.1 \pm 0.6}$\\
5 &$26.1 \pm 1.2$ & $26.1 \pm 0.7$ & $31.2\pm1.5$ & $\mathbf{32.8 \pm 0.2}$\\
\midrule
\text{\tiny Avg horizon len}&$2.71 \pm 0.07 $ & $2.73 \pm 0.04$ & $2.86 \pm 0.11$  &$\mathbf{2.93 \pm 0.04}$\\
\bottomrule
\hline
\end{tabular}
\end{table}

\noindent\textbf{Quantitative Results.} As shown in Table~\ref{method-baselin-result}, NVSPolicy consistently outperforms all recent methods across all metrics on long-horizon tasks. Notably, compared to state-of-the-art, NVSPolicy improves the success rate for completing three consecutive tasks from 51.5\% to 56.3\% and also increases the average successful sequence length. These results confirm its effectiveness in handling both short-horizon and long-horizon sequential tasks under language conditioning.

\noindent\textbf{Qualitative Results.} 
Fig.~\ref{fig:result} presents qualitative comparisons of NVSPolicy with several recent methods in various challenging scenarios. These include scenes with occlusion (a), similar object shape (b), and long-horizon manipulation (c). The results demonstrate that NVSPolicy accomplishes all the manipulation tasks, while the other methods fails.

\subsection{Ablation Studies}
\noindent\textbf{Novel-view Synthesis.}
The novel-view synthesis generates new images from previously unseen viewpoints, providing more context-enhanced observations. To evaluate its contribution, we conduct an ablation study in which the synthesized novel-view image is replaced with an all-zero placeholder (w/o NS). As reported in Table~\ref{method-result}, the performance drops noticeably without this component, particularly on long-horizon tasks. This verifies the importance of novel-view synthesis.

\noindent\textbf{Disentangled Feature Representation.}
Disentangling the visual feature is essential to NVSPolicy in terms of both the meta-skill selection and action estimation. To assess its importance, we perform an ablation study by removing the feature disentanglement module and instead using a simple three-layer convolutional encoder to directly extract the visual feature (w/o FD). As shown in Table~\ref{method-result}, eliminating the feature disentanglement leads to a reduction in the average successful sequence length from 2.93 to 2.74, indicating the necessity of the disentangled feature representation.

\begin{table}[!t]
\caption{Comparisons on the real robot platform. The evaluation metric is the success rate.
}
\label{method-result-real}
\vskip 0.05in
\centering
\small
\setlength{\tabcolsep}{10pt}
\begin{tabular}{ccc}
\hline
\toprule
\diagbox{\textbf{Task}}{\textbf{Method}}& \multicolumn{1}{c}{\textbf {SPIL}~\cite{zhou2024language}} & 
\multicolumn{1}{c}{\bf Ours} \\
\midrule
close the drawer & $10\%$ & $\mathbf{60\%}$\\
push the slider to the right & $20\%$ & $\mathbf{50\%}$\\
lift the blue block from the drawer & $10\%$ & $\mathbf{40\%}$\\
lift the red block from the slider & $20\%$ & $\mathbf{50\%}$\\
push the pink block right & $0\%$ & $\mathbf{20\%}$\\
\midrule
$\text{Avg  success rate}$ & $8\%$ & $\mathbf{44\%}$\\
\bottomrule
\hline
\end{tabular}
\end{table}

\noindent\textbf{Policy Distillation.}
The policy distillation reduces model complexity during inference. To assess its impact, we conduct an ablation study by removing the distillation process and reverting to GenWarp for novel-view image generation (w/o PD). As shown in Table~\ref{method-result}, this results in a slight performance drop, indicating that the policy distillation could not only reduce the inference computational consumption but also increase the robustness of the policy learning.


\begin{figure}[t!]
\centering
    \centering
    \includegraphics[width=1\linewidth]{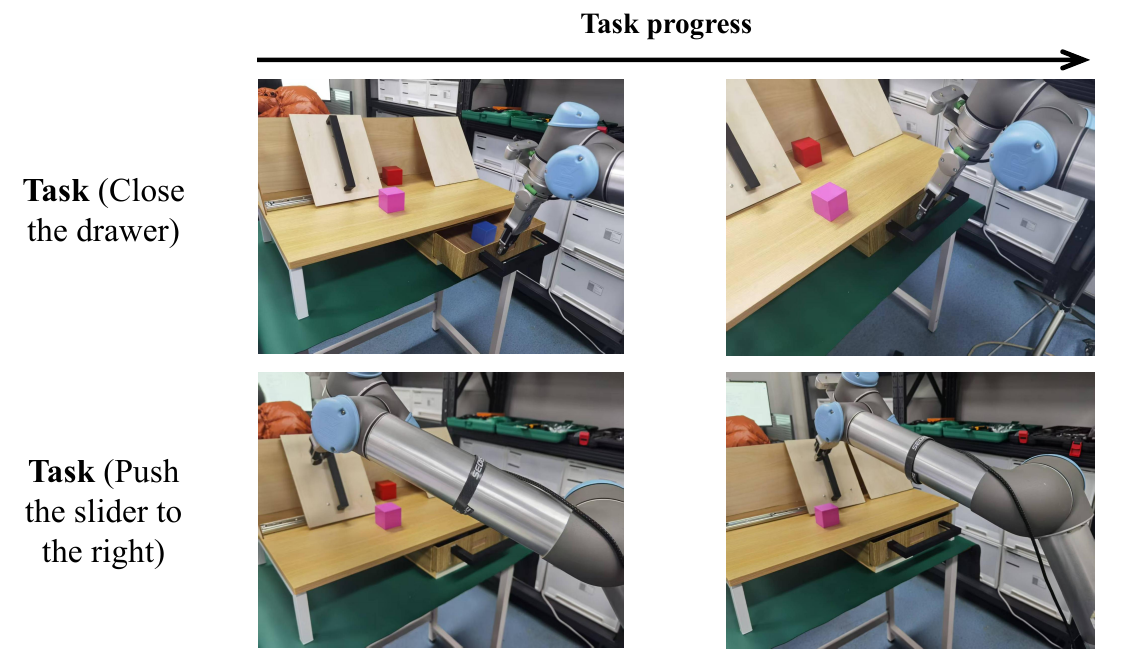}
    \caption{The proposed NVSPolicy shows satisfactory performance on the real robot platform.}
    \label{fig:result_real}
\end{figure}

\noindent\textbf{Adaptive Viewpoint Selection.}
We conduct a quantitative evaluation to assess the impact of our adaptive viewpoint selection strategy, as shown in Table~\ref{method-angel-result}. 
Specifically, we replace the adaptive angle in Eq.~\ref{eq:viewpoint} with several fixed angles for comparison. 
The results demonstrate that the adaptive strategy consistently outperforms fixed-angle baselines, highlighting its effectiveness in balancing information gain and synthesis reliability.

\subsection{Performance on Real-world Robotic Platform}
We evaluate NVSPolicy against the SPIL~\cite{zhou2024language} on a real-world robotic platform across 5 representative tasks. Both methods adopt the adaptation module~\cite{wu2024pseudo} to narrow the sim-to-real gap. The methods are required to be tested 10 times on each task. The success rates are reported in Table~\ref{method-result-real}. It shows that NVSPolicy outperforms the other methods, showing its satisfactory sim-to-real robustness. Fig.~\ref{fig:result_real} illustrates the qualitative results of NVSPolicy.

%% file: conclusion.tex
\section{Conclusion}
\label{sec:conclusion}
We propose NVSPolicy, a language-conditioned hierarchical manipulation framework that integrates adaptive novel-view synthesis with a cycle-consistent variational encoder to provide context-rich and reliable perceptual representations.  
Through extensive experiments, NVSPolicy demonstrates strong generalization and achieves state-of-the-art performance on a long-horizon, language-conditioned robotics benchmark.

%% file: appendix.tex
\section{APPENDIX}
\section*{NOTATIONS AND DEFINITIONS}
\subsection{Inputs, Outputs, Features, and Weights}
\begin{itemize}
\item $a$: Action, represents a 7-DoF relative control vector comprising 3D position displacement, 3D orientation change, and a 1D gripper command.
\item $a^*$: Demonstrated action.
\item $d^\text{cam-ori}$: The distance between the camera and the origin which is equal to the average depth of the input image. 
\item $\mathbb{E}$: The expectation operator.
\item $f_\phi$: Cycle-consistent VAE encoder.
\item $f_\omega$: Cycle-consistent VAE decoder.
\item $f_\alpha$: Action estimator.
\item $I^\text{in-hand}$: The RGB images captured by a wrist-mounted gripper camera.
\item $I^\text{to-hand}$: The RGB images captured by a fixed external camera.
\item $I^\text{synthesis}$: The synthesized image from novel viewpoint.
\item $\{I_1, I_2\}$: Two images captured from the same scene with different viewpoints.
\item $\{ \hat{I}_1, \hat{I}_2\}$: The reconstructed images that the decoder takes as input the exchanged semantic latent variables and the original remaining latent variables to generate.
\item $\{I_{t-1}, I_{t}\}$: RGB frames at previous time steps $t-1$ and current step $t$.
\item $l$: Language instruction.
\item $\mathcal{L}_\text{forward}$: The loss function of cycle-consistant VAE forward process.
\item $\mathcal{L}_\text{reverse}$: The loss function of cycle-consistant VAE reverse process.
\item $\mathcal{L}_\text{KD}$: The distillation loss between $m$ and $n$.
\item $m$: Softened distribution predicted by the student model.
\item $\mathcal{MI}(\cdot)$: The mutual information between two frames.
\item $n$: Softened distribution predicted by the teacher model.
\item $o$: Categorical distribution over the predefined meta-skills.
\item $o^\text{optimal}$: The selected optimal meta-skill.
\item $p(s)$: The prior distributions of semantic latent variable.
\item $p(r)$: The prior distributions of remaining latent variable.
\item $p(o|a)$:  Meta-skill rule-based labels.
\item $p(o)$: Uniform prior distribution.
\item $p(I_{t-1}, I_{t})$: The joint probability distribution of pixel intensities estimated from the co-occurrence histogram of two frames.
\item $p(I_{t-1}), p(I_{t})$: The corresponding marginal distributions.
\item $q_\xi(o|\cdot)$: Categorical distribution of meta-skill selector output.
\item $r$: The remaining latent variables.
\item $\{r^\text{in-hand}, r^\text{to-hand}, r^\text{synthesis}\}$: The remaining latent variables of $\{I^\text{in-hand}, I^\text{to-hand}, I^\text{synthesis}\}$
\item $s$: The semantic latent variables.
\item $s^*$: A semantic latent variable that is sampled from the standard Gaussian distribution.
\item $\{s^\text{in-hand}, s^\text{to-hand}, s^\text{synthesis}\}$: The semantic latent variables of $\{I^\text{in-hand}, I^\text{to-hand}, I^\text{synthesis}\}$
\item $s^\text{synthesis}_\text{teacher}$: The semantic feature vector predicted by teacher model.
\item $s^\text{synthesis}_\text{student}$: The semantic feature vector predicted by student model.
\item $s^{\text{all}}$: Semantic features from all three views.
\item $z^\text{language}$: The language instruction feature embedding.
\item $\theta$: Sample the novel viewpoint whose angle difference in the spherical coordinate to the camera.
\item $\Psi$: A set of image pairs, where each pair contains two images captured from the same scene with different viewpoints.
\item $\psi$: A set of image pairs in $\Psi$.
\item $\pi$: The policy of the robotic manipulation.
\end{itemize}
\subsection{Parameter Settings}
\begin{itemize}
    \item $\omega_1, \omega_2$: The hyper-parameter of viewpoint selection strategy.
    \item $\lambda$: The hyper-paramete to balance $\mathcal{L}_\text{forward}$ and $\mathcal{L}_\text{reverse}$.
    \item $T$: The length of the rolling buffer that maintain recent semantic feature.
    \item $K$: The number semantic feature dimensions.
    \item $\tau$: The temperature parameter that controls the smoothness of the output
    \item $\gamma_1, \gamma_2$: The hyper-parameter of hierarchical policy loss founction.
\end{itemize}